\documentclass{article}
\usepackage{ijcai20}

\usepackage{times}  
\usepackage{helvet}  
\usepackage{courier}  
\usepackage{url}  
\usepackage{graphicx}  
\usepackage{amsfonts}
\usepackage{amsmath}
\usepackage{amssymb}
\usepackage{mathptmx}
\usepackage{nicefrac}
\usepackage{booktabs} 
\usepackage{makecell} 
\usepackage{graphicx}
\usepackage[font=small]{subcaption}
\usepackage[font=small]{caption}
\usepackage{xargs} 
\usepackage{multicol}
\usepackage{booktabs}
\usepackage{dirtytalk}
\usepackage[most]{tcolorbox}

\usepackage{color}
\usepackage[textsize=scriptsize]{todonotes}
\definecolor{blue}{RGB}{0, 93, 170}	
\definecolor{darkgreen}{RGB}{0, 102, 0}

\newcommand{\ignore}[1]{}
\newcommand{\typewriter}[1]{\textsc{#1}}
\newcommand{\tabletype}[1]{\textsc{#1}}
\newcommand{\tabletypesmall}[1]{\texttt{#1}}

\definecolor{block-gray}{gray}{0.95}
\newtcolorbox{blockquote}{colback=block-gray,grow to right by=-1mm,grow to left by=-1mm,boxrule=0pt,boxsep=0pt,breakable}

\usepackage{xspace}

\newcommand{\citet}[1]{\citeauthor{#1}~\shortcite{#1}}
\newcommand{\citep}{\cite}

\begin{document}
%


\title{Path-Based Contextualization of Knowledge Graphs for Textual Entailment}

\author{
Kshitij Fadnis$^\dag$ \and
Kartik Talamadupula$^\dag$\and
Pavan Kapanipathi$^\dag$ \\
Haque Ishfaq$^\S$ \and 
Salim Roukos$^\dag$ \and
Achille Fokoue$^\dag$
\affiliations
$^\dag$IBM Research\\
$^\S$McGill University
\emails
\{kpfadnis,krtalamad,kapanipa,roukos,achille\}@us.ibm.com,
haque.ishfaq@mail.mcgill.ca
}

\maketitle
\begin{abstract}

In this paper, we introduce the problem of {\em knowledge graph contextualization} -- that is, given a specific NLP task, the problem of extracting meaningful and relevant sub-graphs from a given knowledge graph. The task in the case of this paper is the textual entailment problem, and the {\em context} is a relevant sub-graph for an instance of the textual entailment problem -- where given two sentences \typewriter{p} and \typewriter{h}, the entailment relationship between them has to be predicted automatically. We base our methodology on finding paths in a cost-customized external knowledge graph, and building the most relevant sub-graph that connects \typewriter{p} and \typewriter{h}. We show that our path selection mechanism to generate sub-graphs not only reduces noise, but also retrieves meaningful information from large knowledge graphs. Our evaluation shows that using information on entities as well as the relationships between them improves on the performance of purely text-based systems. 




\end{abstract}

\section{Introduction}
\label{sec:introduction}

\begin{figure}[h]
    \centering
    \includegraphics[width=0.8\columnwidth]{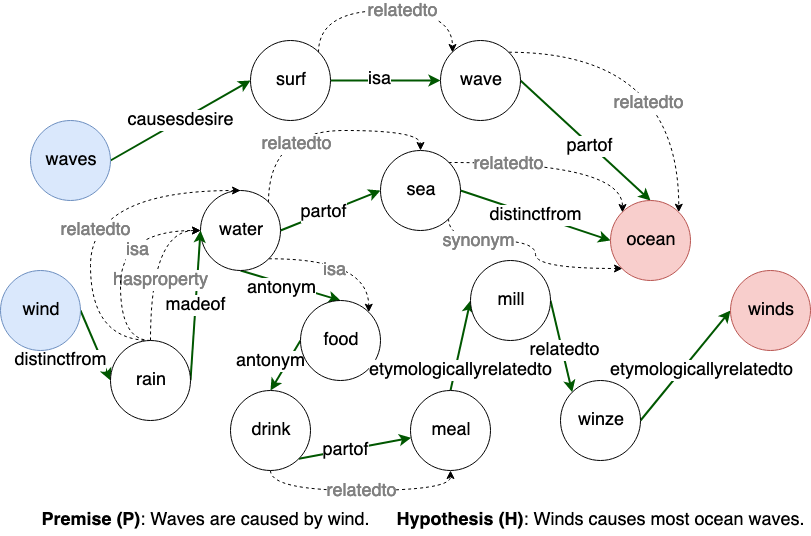}
    \caption{An NLI instance situated in a knowledge graph. Premise nodes are blue, and hypothesis nodes are red.}
    \label{fig:kg_example}
    \vspace{-3mm}
\end{figure}

Textual Entailment or Natural Language Inference (NLI) (both terms used interchangeably) is one of the fundamental tasks in Natural Language Processing (NLP). 
The goal of NLI is: given two sentences, a premise \typewriter{p} and a hypothesis \typewriter{h}, determine whether their relationship is  {\tt entailment/contradiction/neutral} -- the sentences in Figure~\ref{fig:kg_example} depict an instance of the this task. 
NLI has garnered significant attention in the NLP community because of: (1) its usefulness in evaluating the ability of a system to reason (in order to determine entailment); and (2) its use for various downstream applications such as text summarization and question answering (QA). 
NLI tasks are often construed as multi-class text classification problems. Models developed for NLI have largely focused on encoding the textual content of the premise and the hypothesis using various contextual word embeddings, including Glove~\cite{wang2016learning} and embeddings generated from transformer based architectures such as BERT~\cite{devlin2018bert}. Of late, another line of work that has emerged is the augmentation of textual information with external knowledge bases / knowledge graphs (KG) to improve performance on NLI~\cite{kapanipathi2020infusing,wang2019improving,chen2018}. However, external knowledge sources have not been used to their full potential; in this work, we address this by building a model that explicitly harnesses relation (or path) based information from knowledge graphs.


Knowledge graphs contain structured information represented as directed labeled graphs whose nodes denote concepts and edge labels are the relationships between those concepts. Figure~\ref{fig:kg_example} depicts a sub-graph of a commonsense KG called ConceptNet~\cite{speer2017conceptnet}. Specifically, the sub-graph represents the {\em context} for the premise and hypothesis texts. 
However, the process of building an NLI model to harness such sub-graphs is faced with the following challenges: (1) KGs are large and noisy, and hence extracting relevant and meaningful sub-graphs~\cite{kapanipathi2020infusing,wang2019improving} is non-trivial; and (2) effectively encoding all the information contained in the extracted sub-graphs is non-trivial due to the large numbers of entities and relationships that are returned.

In order to address the first challenge -- extracting relevant sub-graphs from noisy knowledge graphs -- we devise a mechanism that selects meaningful paths between concepts mentioned in premise and hypothesis. These paths are derived by computing the shortest paths between premise and hypothesis nodes sets, using different cost functions (heuristics) to predict the closeness between concepts. Each cost function (heuristic) gives rise to a different, cost-customized copy of the KG in the following manner: we keep the structure of the graph unchanged, but add a weight to each edge that is computed using a specific cost function (heuristic). In this way, we invert the traditional notion of the heuristic as used in A$^\star$ search~\cite{hart1968formal}: instead of assigning cost to each node in the graph, we transfer that cost on to each outgoing edge of that node. Next, our NLI model takes {\em both} entities and relationships along these shortest paths as input; thereby augmenting the premise and hypothesis texts with all the information from the sub-graph.\\

\noindent {\bf Contributions}: Our work is the first to encode information from the entire path (entities {\em and} relationships) from KGs for the NLI problem. Our quantitative results show that taking the entirety of the path-based information is more useful than merely entity-based~\cite{wang2019improving,chen2018} or structure-based~\cite{kapanipathi2020infusing} information alone. We thus show that harnessing the full path -- specifically relationships -- does quantifiably improve the performance of NLI systems. Our path selection mechanism also addresses the issue of noise from KGs by introducing: (1) the choice of multiple automatically computed heuristics; and (2) the path length to be considered -- as parameters that can be tuned. 


\section{Related Work}
\label{sec:related_work}



\subsection{Natural Language Inference}
\label{subsec:related_nli}

The introduction of large-scale NLI datasets~\cite{bowman2015large,khot2018scitail} has led to increase in the number of supervised classification models that are being invented for the NLI task. \citet{bowman2015large} proposed an LSTM-based neural network model which was the first generic neural model without any hand-crafted features.  
``Matching aggregation'' approaches, on the other hand, exploit various matching methods to obtain an interactive premise and hypothesis space. For example, \citet{wang2016learning} perform a word-by-word matching of the hypothesis with the premise using match-LSTM (mLSTM). \citet{rocktaschel2015reasoning} use a weighted attention mechanism to get an embedding of the hypothesis conditioned on the premise. \citet{parikh2016decomposable} decompose the entailment problem into sub-problems through an intra-sentence attention mechanism, and are thus able to parallelize the training process. \citet{ghaeini2018dr} encode both the premise and the hypothesis conditioned on each other using BiLSTM, and then use a soft attention mechanism over those encodings. Recently, transformer architectures such as BERT \cite{devlin2018bert} and RoBERTa \cite{liu2019roberta} have significantly improved the performance of NLI systems, particularly on leaderboards~\cite{zhang2018know,liu2019multi}. These models require a lot of training data and are computationally intensive. Due to these drawbacks and the fact that our focus is on exploring the use of knowledge graphs (rather than optimizing solely text-based models), we use match-lstm (mLSTM)~\cite{wang2016learning}, which is a simple and effective text-based model.  

\subsection{Knowledge Graphs and NLI}
\label{subsec:related_kg}

Although there have been extensive studies on the NLI task, the potential for exploiting external knowledge encoded in knowledge graphs (KGs) has not been explored in enough detail. Among the few existing approaches, \citet{chen2018} use WordNet as the external knowledge source for NLI. They generate features based on WordNet using the relationships in it. However, WordNet -- being a lexical database -- possesses very few linguistic relationships among entities, and thus its richness as an external knowledge source is limited. There are other KGs such as DBpedia~\cite{auer2007dbpedia} and ConceptNet~\cite{liu2004conceptnet,speer2017conceptnet} that have become popular due to the richer information contained in them. One issue with expressive KGs such as DBpedia and ConceptNet  is that they are quite massive in terms of the nodes and edges contained in them, which makes it hard to extract relevant information.

The closest approaches to our current work are those of \citet{wang2019improving,kapanipathi2020infusing,chen2018}. KIM~\cite{chen2018} is one of the first models to use features that are engineered using WordNet as the KG. This approach has a few drawbacks: (1) the engineered features are specific to the KG and the model is not flexible enough to augment any text-based models; and (2) Wordnet is primarily a linguistic knowledge base with $16$ distinct relations and approximately $160,000$ concepts. Such a small KG is restrictive (in terms of the domain) in comparison to KGs such as ConceptNet that contain more than a million concepts and more than 50 distinct relationships. ConSeqNet~\cite{wang2019improving} and KES~\cite{kapanipathi2020infusing} address the drawbacks of KIM by using ConceptNet and a modular approach to incorporate knowledge graphs into text-based models. 
However, both models do not effectively use all the information from the KG (entities and relationships). While ConSeqNet uses only entities in the knowledge graph, KES introduces graph convolutional networks (GCNs) to capture the structure of the sub-graphs (including the entity information). However, KES uses only 3 relationships (edge, inverse, and self-loop). These relationships are not present in ConceptNet; but there are 50 other distinct relationships that can be utilized. Our model, on the other hand, takes as input paths that include both entities and relationships. 

\section{Methodology}
\label{sec:methodology}

The goal of our approach is 
to identify paths between premise and hypothesis concepts to augment the text content of the entailment task.  
To do this, first, we create different versions of the ConceptNet knowledge graph that feature customized costs as the weights on the relationship-edges -- we call these {\em customized cost graphs}. Following this, for each labeled premise and hypothesis pair in the 
dataset of interest, we extract the concepts from each respective sentence. We then take the Cartesian product of the premise and hypothesis concepts (respectively) to create ordered premise-hypothesis entity pairs; and then find the shortest path between each of these entity pairs in the customized cost graphs. For each premise-hypothesis pair (a textual entailment problem instance), the collection of shortest paths thus found is then associated with the corresponding label (entails/contradicts/neutral) for purposes of learning  to classify accurately (described in more detail in Section~\ref{sec:graph_learning}). 




\subsection{Customized Cost Graphs}
\label{subsec:customized_cost_graphs}

The first step towards constructing meaningful sub-graphs as context is to select the external knowledge source. In this paper, we selected  the ConceptNet~\cite{speer2017conceptnet} graph, which contains crowdsourced and expert-created commonsense knowledge. 
Next, in our quest to retrieve the right knowledge (sub-graph) from the KG, 
we follow a path-based approach. Given concepts mentioned in premise and hypothesis text, we generate paths between them. In order to generate paths, we weight the relations in the KG which facilitates computing shortest paths between concepts. 
Determining the  weights for the relations is grounded in a simple insight: not all relations between concepts are equal. Put another way, the ConceptNet graph -- which is made up of concepts and the relation edges that connect them -- needs to be weighted in order to reflect this fact. Therefore, we create copies of the ConceptNet knowledge graph with customized weights on the relation-edges.
This is achieved by 
treating the weights on the graph edges as a {\em cost} that is incurred any time that specific edge has to be traversed. 
We call each of the copies of ConceptNet thus produced a {\em cost graph}, and demonstrate the use of these various cost graphs in Section~\ref{subsec:shortest_paths}. In the following, we detail three heuristics that we use to generate the edge costs and eventually the cost graphs.


\subsubsection{Default Cost (DC)}

This is the simplest case we consider, where we assign every single edge in our target graph (ConceptNet) a cost of $1.0$. This essentially turns the path-finding problem between two given nodes on the graph into a problem of minimizing the number of hops: the shortest hops give the most efficient path.



\subsubsection{Relation Frequency (RF)}

This heuristic seeks to automate the computation of the edge weight, and base that computation on some feature of the graph itself. The first such idea is to simply count the frequency of the relations associated to a concept. We specifically implement this heuristic as the normalized count of the number of outgoing edges bearing the same relation name from a given node. That is, given a node $n$ that represents a concept in the graph, and $rel(n)$ the set of outgoing edges from $n$, we represent the cost $c_i$ for an edge $e_i \in rel(n)$ as $c_i = \dfrac{|e_i| }{| rel(n) |}$. For example, consider a node $n_1$ that has three outgoing edges: $\{ e_1, e_2, e_1 \}$. Using the above formula, the weights of the $e_1$ edges would be set to $c_{e_1} = 0.67$, while the edge $e_2$ would have a cost of $0.33$. This ensures that the relation that is ``rarer'' is given a lower cost, and is favored by a shortest-path algorithm if there is more than one way to travel from a node to its neighbor. 

\subsubsection{Global Relation Frequency (GRF)}

The final heuristic that we consider builds on top of the relation frequency metric by addressing a significant issue: the presence of common relations that occur throughout the knowledge graph, but may occur relatively fewer times at any one individual node. An example of such a relation is \typewriter{Is-A}; while this relation is likely to occur relatively fewer times at any given node, it is clear that it occurs throughout the graph. We want to ensure that a truly rare relation that participates in an entailment instance is thus given more importance (and subsequently less cost) than one which occurs throughout the graph. To do this, we follow the inspiration of TF-IDF, which is often used to address similar issues in text corpora.

We first compute the Inverse Node Frequency (INF) (the analog of IDF) for every relation in the graph. Given a graph with node-set $N$, let the quantity $n_{rel_i}$
be the number of 
nodes that feature $rel_i$ as an outgoing edge. The INF for edges with the relation label $rel_i$ can then be calculated as $INF_{rel_{i}} = \log{\dfrac{| N |}{n_{rel_i}}}$. Next, we compute the normalized Relation Frequency (RF) as in the previous section. Thus given a node $n \in N$ with a set of outgoing edges $e$, the RF for an edge with relation $i$ can be calculated as $RF_{rel_{i}} = \dfrac{| e_i |}{| rel(n) |}$. Since we are interested in promoting ``rarer'' relations by associating lower cost with them, we invert INF during the the calculation of the final cost metric, giving us the cost as $c_i = RF_{i} \times \dfrac{1}{INF_{i}}$.

\subsection{Ordered Premise \& Hypothesis Pairs}
\label{subsec:ordered_subsets}

Once we generate the various cost graphs as described above, we use those respective graphs to classify the two sentences in a given textual entailment instance. As before, let us assume that this instance is denoted $\tau = \langle p, h \rangle$, where $p$ is the premise sentence and $h$ is the hypothesis sentence. The first step we take is to represent each sentence based on the concepts mentioned in them: that is, we collapse the representation of a sentence into an ordered set of concepts from the sentence that also appear in ConceptNet.
Let us denote these ordered sets as \typewriter{p} and \typewriter{h} respectively. Since we do not know which concepts in the premise and which ones in the hypothesis contribute directly to the classification of the entailment relationship, we take the cartesian product of the two ordered sets \typewriter{p} and \typewriter{h} to generate the set of all possible ordered pairs between $p$ and $h$. This set $S = \typewriter{p} \times \typewriter{h} = \{ (a, b)\ | \ a \in \typewriter{p},\ b \in \typewriter{h} \}$ is then used as the input for the shortest path generation step.


\subsection{Shortest Paths}
\label{subsec:shortest_paths}

Once we have the sets of premise-hypothesis concept pairs from Section~\ref{subsec:ordered_subsets}, we move on to finding all shortest paths between the concepts of each pair, for every cost graph outlined previously. We employ {\tt NetworkX}'s ~\cite{hagberg2008exploring} implementation of the Dijkstra shortest-path algorithm. Since ConceptNet has about 1 million nodes and well over 3 million edges, finding shortest paths is a computationally expensive process. 
Additionally, after an analysis of concept pairs from ConceptNet that feature more than one direct edge between them (multi-edges), we find that the most common relationship ({\tt RelatedTo}) occurs about $83\%$ of the time. The second most common relationship ({\tt FormOf}) occurs in about $33\%$ of cases. Further, these two relations co-occur around $30\%$ of the time, and of those cases, for about $97\%$ of the time, they are the {\em only} two relations connecting that concept pair. All of these support our hypothesis that selecting at random between paths that contains either of these relationships will not have a significant impact on the NLI classification problem. We use this as motivation to reduce the problem from one of finding all possible shortest paths between premise-hypothesis concept pairs, to one of finding a {\em single} shortest path. 

\section{Using KG Information via Shortest Paths}
\label{sec:graph_learning}

\begin{figure}[t]
    \centering
    \includegraphics[width=\columnwidth]{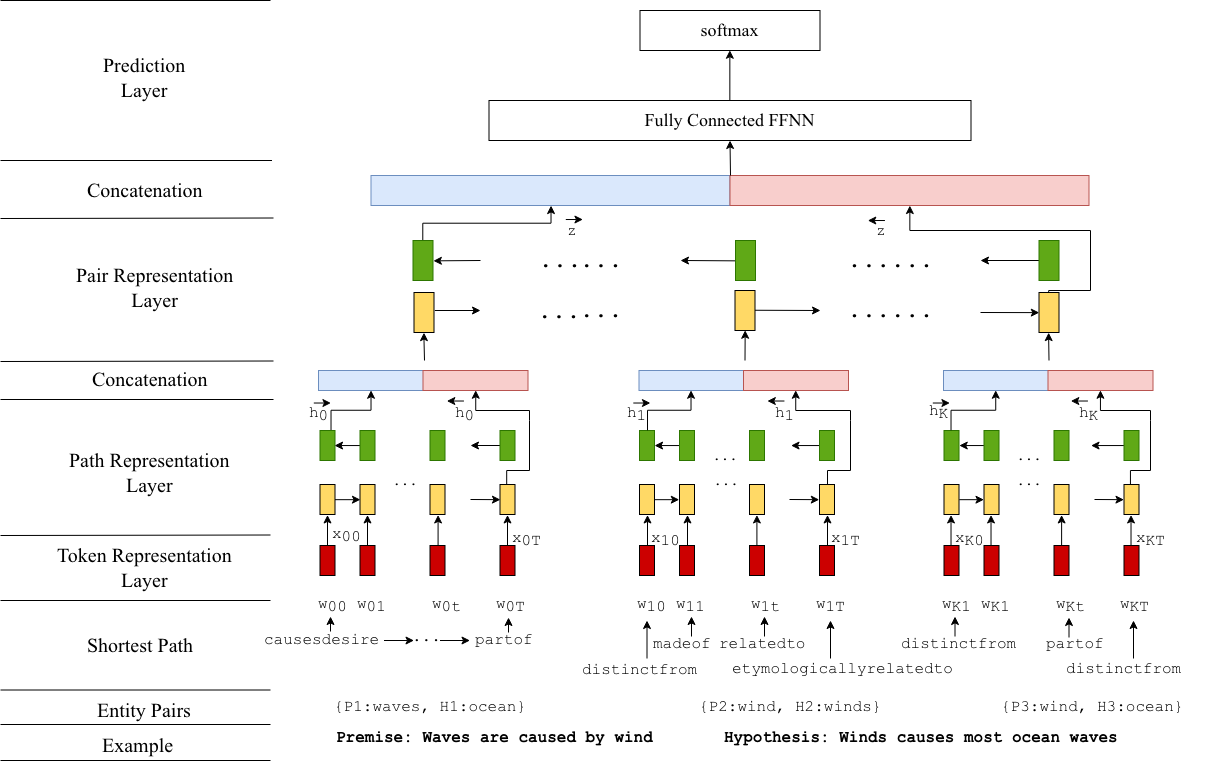}
    \caption{Architecture of our GRN model.}
    \label{fig:rnn}
    \vspace{-2mm}
\end{figure}

Once the pairwise shortest paths are generated, we need to use them in a way that enables us to train on labeled textual entailment instances, in order to make predictions on new instances. In this we focus particularly on the {\em path} part of the shortest paths -- that is, we are interested in considering the relations used to connect a given premise and hypothesis pair. This harks back to our hypothesis in Section~\ref{sec:methodology} that the relationships between concepts in the textual entailment instance are key to identifying the overall entailment relationship. In this section, we detail the procedure by which we use the specific sequence order in which relationships appear in the shortest path. This approach is in contrast to the work of \citet{wang2019improving}, which only considers entity-level information and completely ignores relationships.

\subsection{Text Model: mLSTM}
\label{subsec:matchLSTM}

Most models for the NLI problem use only the premise and hypothesis sentence as input; due to this fact, we decided to use match-LSTM (mLSTM) \cite{wang2016learning} as our text-based model. The specific implementation of mLSTM that we use encodes both premise and hypothesis as Bi-GRUs (as against Bi-LSTMs), and a fixed representation of the hypothesis that is premise-attended is output. 

\subsection{Recurrent Neural Networks}
\label{subsec:rnn}

In order to model the path information, we make use of the sequentiality inherent in a shortest path. Recent work on Graph Convolutional Recurrent Networks (GCRN) ~\cite{seo2018structured} has explored representing sequential graphical structures as fixed representations. One of the major difference between that approach and the one we take in this work is the degree or {\em level} of sequentiality. In our current problem, we are faced with two levels of sequential information. One of these is at the level of ordered premise-hypothesis entity pairs. The other is at the level of the path, which is represented as a sequence of relations, entities, or both; per premise-hypothesis entity pair. 




We first describe how we process the shortest paths to capture the bi-level sequentiality inherent in them. As before, we assume each textual-entailment instance $\tau$ consists of premise ($p$) and hypothesis ($h$), which together constitute a sentence pair. After processing each $\tau$ as outlined in Sections~\ref{subsec:ordered_subsets} and \ref{subsec:shortest_paths}, we obtain an ordered set of shortest paths. Each of these shortest paths can be represented by either the concepts along that path (alone), the relations along that path (alone), or a combination of the concepts and relations both. Our work follows various hierarchical architectures that have been proposed for different learning-centric  tasks~\cite{sordoni2015hierarchical,li2015hierarchical}.
The hierarchical assumption formulates a sequence at two levels: (1) a sequence of tokens for each pair; and (2) a sequence of pairs. We model this as two recurrent neural networks.

Figure~\ref{fig:rnn} shows the architecture of our {\em Graph Recurrent Network} (GRN) architecture. We describe the functioning of the GRN via a simplified working example. Consider the two sentences: \typewriter{Waves are caused by wind} (premise); and \typewriter{Winds causes most ocean waves} (hypothesis). As described in Section~\ref{subsec:ordered_subsets}, we first find all possible premise-hypothesis concept pairs. This particular example gives us $12$ such pairs: 3 premise (\typewriter{waves, caused, wind}) times 4 hypothesis (\typewriter{winds, causes, ocean, waves}) entities. We further simplify for the sake of exposition and focus on three entity pairs: (\typewriter{waves}, \typewriter{ocean}), (\typewriter{wind}, \typewriter{winds}), and (\typewriter{wind}, \typewriter{ocean}). As explained in Section~\ref{subsec:shortest_paths}, we identify shortest paths for each of these pairs. For example, for the pair (\typewriter{waves}, \typewriter{ocean}), the shortest path looks like: \typewriter{waves $\rightarrow$ causesdesire $\rightarrow$ surf $\rightarrow$ isa $\rightarrow$ wave $\rightarrow$ partof $\rightarrow$ ocean}, where \typewriter{waves}, \typewriter{surf}, \typewriter{waves} and \typewriter{ocean} are entities along the path; and \typewriter{causesdesire}, \typewriter{isa} and \typewriter{partof} are the relationships connecting them in sequential order. 

The GRN  model can take either relations, concepts, or relations plus concepts as its input. In Figure~\ref{fig:rnn}, we show an instance where relations are fed as input to the token-representation layer. At this point, the tokens -- which are relations in this case -- are transformed into vector representation using an embedding matrix. The transformed representations are then fed to a bidirectional Recurrent Neural network (RNN) in the sequence order captured by the shortest path. The final hidden states from the bidirectional RNN are then concatenated to form a representation for the whole path. Thus after passing through the path representation layers, we have vector representations for each of the entity pairs. These representations are then fed into a second bidrectional RNN in the order prescribed by the ordered set of entity pairs. Once the final hidden states of the pair-level encoder are concatenated, a feed-forward network with rectified linear units (ReLU) and linear activation with softmax layer is used as a final prediction layer.

\begin{table*}[t]
\centering
\resizebox{\linewidth}{!}{
\begin{tabular}{c|ccc|ccc|ccc|c} 
\toprule
& \multicolumn{3}{c|}{{\bf \tabletypesmall{SciTail}}}  & \multicolumn{3}{c|}{{\bf \tabletypesmall{SNLI}}} & \multicolumn{3}{c}{{\bf \tabletypesmall{BreakingNLI}}} & Average  \\
 & \tabletype{DC} & \tabletype{RF} & \tabletype{GRF} & \tabletype{DC}  & \tabletype{RF} & \tabletype{GRF} & \tabletype{DC} & \tabletype{RF} & \tabletype{GRF} & Improvement                 \\
\midrule
   Relations  & 85.60 (3) & 84.95 (3) & 85.61 (3) & 84.31 (3) & 84.50 (4)  & 84.38 (3) & 65.08 (3*) & 65.40 (4*) & 73.97 (3*) & 2.2         \\
   Entities  &	84.24 (2) &	84.15 (5) &	84.71 (3) &	84.17 (3) & 84.43 (5) & 84.60 (4) & 68.23 (3*) & 67.90 (5*) & 67.30 (4*) & 1.7 \\
   Relations + Entities  &	84.81 (4) &	85.37 (4) &	85.42 (2) & 84.91 (2) & 84.00 (5) & 84.75 (4) & 69.69 (2*) & 70.39 (5*) & 67.41 (4*) & 2.6\\
   \hline
   Text Only (mLSTM)& \multicolumn{3}{c|}{{ 82.54}}  & \multicolumn{3}{c|}{{ 83.60}} & \multicolumn{3}{c}{{ 65.11}} \\
\bottomrule
\end{tabular}
}
\caption{Accuracy values (in $\%$) for GRN + mLSTM experiments with input being relations, entities, and relationship + entities. * Parameters are the same from the SNLI tuned model. 
}
\label{tab:grn_experiments}
\end{table*}

\begin{table*}[t]
\centering
\resizebox{\linewidth}{!}{
\begin{tabular}{lccc} 
\toprule \\
 & \tabletype{SciTail} & \tabletype{SNLI} & \tabletype{BreakingNLI}\\
\midrule
   KIM (Entity-based Feature Engineering)~\cite{chen2018} & - & \textbf{88.6}* & \textbf{87.42}* \\
   ConSeqNet (Entities only)~\cite{wang2019improving}  & 85.2* & 83.34* & 61.12* \\
   KES (Entities only + One hop neighborhood structure)~\cite{kapanipathi2020infusing} &	82.22* &	83.94* &	\textbf{78.72}* \\
   GRN + mLSTM (Paths including Entities and Relationships) [This Paper]  &	\textbf{85.37} (RF-4) &	\textbf{84.91} (DC-2) & 69.69 (DC-2) \\
\bottomrule
\end{tabular}
}
\caption{Comparison with State of the Art Knowledge Based NLI techniques.}
\label{tab:soa_comparison}
\end{table*}

\subsubsection{Token-level Encodings}
\label{subsubsec: token_encoder}

Each pair $pair_i$ consists of a sequence of tokens $w_{it}, t \in [0, T]$ which are embedded using an embedding matrix $W_t$ as $x_{it} = W_t w_{it}$. Then the bidirectional token-level RNN -- a GRU in our case -- is used to form a fixed length representation by concatenating the final state  from forward($\overrightarrow{h_{it}} = \overrightarrow{GRU}(x_{it}), t \in [1, T]$) and backward ($\overleftarrow{h_{it}} = \overleftarrow{GRU}(x_{it}), t \in [T, 1]$) passes in the GRU. This yields $h_i = [\overrightarrow{h_{iT}}, \overleftarrow{h_{i0}}]$. Note that we use ComplEx  ~\cite{trouillon2016complex} knowledge graph embeddings for token-level embeddings. These emebeddings are trained on ConceptNet using OpenKE (\url{https://github.com/thunlp/OpenKE}).


\subsubsection{Pair-level Encodings}

The input to the pair-level encoder is a sequence of token-level representations $h_1, h_2, \dots, h_K$. Then, a bidirectional GRU computes the fixed length representation as: $\overrightarrow{z} = \overrightarrow{GRU}(h_{k}), t \in [1, K]$; $\overleftarrow{z} = \overleftarrow{GRU}(h_{k}), k \in [K, 1]$; $z = [\overrightarrow{z}, \overleftarrow{z}]$.


\section{Experimental Setup}
\label{sec:experimental_setup}


In this section, to bolster reproducibility, we talk about our experimental setup: this includes the dataset, the knowledge graph used,  various initializations, and hyperparameters. 

\subsection{Dataset \& Knowledge Graph}
\label{subsec:setup_dataset_kg}

{\bf Datasets}: We use multiple NLI datasets to evaluate our approach: SNLI~\cite{bowman2015large},  SciTail~\cite{khot2018scitail},  and BreakingNLI~\cite{glockner2018breaking}. SNLI is a manually labelled, crowdsourced dataset for textual entailment. It is one of the largest NLI datasets (500K sentence pairs), and the scale of this dataset has been one of the major factors in the increase in learning-based models for NLI. SciTail is a science domain entailment dataset (\~30K sentence pairs), created from a corpus of science domain multiple choice questions for $4$th and $8$th grades. While crowdsourcing is useful in creating large labelled datasets, a prominent drawback is the introduction of artifacts that can be easily captured and exploited by learning systems. In order to address this issue in SNLI, ~\citet{glockner2018breaking} created a new test set called BreakingNLI (BNLI). Specifically, BreakingNLI is an adversarial dataset created using liguistic knowledge sources.   

\vspace{2mm}
\noindent {\bf Knowledge Graphs}: There are multiple open knowledge sources available such as DBpedia \cite{auer2007dbpedia}, WordNet \cite{miller1995wordnet}, and ConceptNet \cite{speer2017conceptnet}. Each knowledge source contains different kinds of information, and 
selecting the right knowledge source for a specific task or dataset is non-trivial. \citet{wang2019improving,kapanipathi2020infusing}'s work provides some guidance on this task, by evaluating the relevance of each of these knowledge bases to the SciTail dataset; their conclusion is that ConceptNet is the best KG for NLI datasets.




\subsection{Initializations \& Hyperparameters}

In the Graph Recurrent Network (GRN) model, we used ComplEx ~\cite{trouillon2016complex} 
embeddings for the token-level encoder, with the embedding dimension set to $300$. The token-level and pair-level encoders used single-layered bidirectional GRUs with a hidden size of $300$. Parameters were not shared between token-level and pair-level encoders. A two-layered fully-connected feed-forward neural network with ReLU and linear activation, and dropout of $0.2$ and $0.0$ respectively, was used for the prediction layer. The size of the hidden layer was set to $200$. 

In order to reduce the noise from the KG, we developed our method for path-based extraction of sub-graphs. However, our first set of results using all paths between premise and hypothesis concepts (hops: min $2$ - max $21$) did not show any significant improvement in performance; we omitted these experiments due to space constraints. Thereafter, we added another hyperparameter to our model: the maximum path length (ranging between $2$-$5$) in the contextual sub-graph. This is a hyperparameter tuned for each of our models. 

Our models were implemented with {\tt AllenNLP}, a popular NLP library. We tuned the hyperparameters for the models using the validation set. We used a sigmoid function and minimized cross-entropy loss for training and updating the model. The training cycle involved a $150$ epoch run, with a $20$ epoch patience cutoff. The batch size was set to $64$, and gradients were clipped at $5.0$. The trainer was configured to use the Adam optimizer with a learning rate of $0.001$. 







\section{Results}
\label{sec:results}

In this section, we outline our results and present our viewpoint on the quality of the datasets and knowledge graphs that impact the results. Keeping with our main claims, we first evaluate different heuristics for generating the paths, and show that including both entities and relationships in our path-based model improves the performance of the text-based system. Next, we also show that our KG-based approach that utilizes both relationships and entities from KGs is comparable to the current state of the art.

Table~\ref{tab:grn_experiments} shows an ablation study of our system that augments mLSTM (text-based model) with GRN (path-based model). The reported results have the path-length hyper-parameter tuned (number in  parenthesis beside each accuracy value in Table~\ref{tab:grn_experiments}). In this study, we have primarily focused on two dimensions: (1) Input, which includes concepts, relations, and relationship+concepts; and (2) the three heuristics used for generating customized cost graphs. First, we see that all models trained using our path-based approach perform better than mLSTM (absolute increase between $0.6\%$-$8\%$). Furthermore, the average increase in performance is the best when both relations and concepts are used in the path ($2.6$). A significant portion of this improvement can be attributed to the introduction of relations into the model, since using relations alone shows an average increase of $2.2\%$ over text-based models. Along the dimension of the heuristics, in most cases, {\tt GRF} performs slightly better than the other two cost functions. The results thus show no significant difference between the heuristics; however, the primary purpose of our current work is to recommend the use of cost functions and not to plump for any specific heuristic.  

Table~\ref{tab:soa_comparison} compares our approach with the state of the art in KG-based approaches. KIM performs the best on SNLI and BreakingNLI. It is important to note that KIM's approach tightly integrates manually engineered features from WordNet with the text-based models; this makes it non-trivial to scale and generalize KIM to state of the art text-based models or other knowledge graphs. ConSeqNet, KES, and our GRN approach address this drawback by utilizing ConceptNet, a larger and much sophisticated commonsense KG. Unfortunately, this generalizability comes with a minor compromise on the performance on SNLI and BreakingNLI.
Using this generalizability as common grounds for comparison, our path-based GRN model performs better than both KES and ConSeqNet on SciTail and SNLI; but not on BreakingNLI (more details in Section~\ref{sec:discussion}). Furthermore, the path-based approach is the only approach that effectively uses information from both relations and concepts in the KG. ConSeqNet uses only concepts; and KES uses the structure of that subset of the graph which is one-hop away from premise and hypothesis concepts, ignoring all relations. 


Finally, perhaps the most exciting result pertains to the reduction of noise and the extraction of relevant knowledge from a KG, which is the primary goal of this work (c.f. Table~\ref{tab:soa_comparison}). On the SciTail dataset, where our method shows the largest improvement in performance over KES, we also extracted the average number of entities and relations for each method. GRN (with \typewriter{RF} heuristic, length $4$) brings in an average of $13.72$ entities and $10.24$ relations; in comparison to $33.27$ entities and $22.09$ edges for KES -- this for a net performance improvement of $3.15\%$. Similarly, on the SNLI dataset, GRN (with \typewriter{DC} heuristic, length $2$) brings in an average of $6.79$ entities and $4.40$ relations; compared to $16.15$ entities and $25.13$ relations for KES -- resulting in a net improvement of $0.97\%$.  These two results validate our point that our technique improves on the state of the art by offering flexibility in the paths that can be considered (as against KES' rigid limitation of one-hop expansion), and thus dynamically limiting noise in the NLI task.

\section{Discussion}
\label{sec:discussion}

As we previously mentioned in Section~\ref{sec:experimental_setup}, we chose the BreakingNLI dataset to show the robustness of our approach against adversarial perturbation of the concepts in premise sentences. Even though our proposed GRN + mLSTM model improves on text-based models and ConSeqNet~\cite{wang2019improving}, it still significantly underperforms with respect to the KES model~\cite{kapanipathi2020infusing} and KIM model \cite{chen2018}. BreakingNLI is automatically created using a linguistic knowledge base; we surmise that using a WordNet-based model (KIM) on it extends an unfair advantage. This is because WordNet is also a linguistic knowledge base, with similar relations between the same concepts that were used to create the BreakingNLI test set. 

Furthermore, our analysis shows that BreakingNLI by its very design favors replacement concepts which are one-hop away from the concept to be replaced. For example, colors like {\tt red} and {\tt blue} are connected to each other in ConceptNet with the  \typewriter{relatedTo} relation, as well the \typewriter{antonym} relation. The same can be said of many other categories like \typewriter{cardinal}, \typewriter{ordinal}, \typewriter{drinks}, \typewriter{vegetables}, and so on. The KES model in particular  considers the one-hop neighborhood from the concepts mentioned in premise and hypothesis; this provides an unfair advantage during its evaluation on the BreakingNLI dataset. While both KIM and KES benefit from the one-hop replacement design of BreakingNLI, both approaches are less generalizable to other datasets due to manually engineered features, and the single-hop constraint. Our GRN approach, on the other hand, uses paths between multiple hops by including information from both concepts and relationships in the KG. This point is shown at the end of Section~\ref{sec:results}. 


\section{Conclusion}

In this paper, we presented the notion of {\em contextualizing} a knowledge graph by customizing the edge-weights in that graph with costs produced by various heuristic functions. We used these {\em cost customized} graphs to find shortest paths for different instances of the NLI problem, and trained two different classifiers using the sequence information from the shortest paths. Our results over multiple datasets show that our approach -- which extracts and utilizes path-based information from KGs -- is useful in augmenting text-based NLI models. Fully interpreting these paths and ranking them qualitatively based on their value to the NLI task is our future work.

\bibliography{ureqa}
\bibliographystyle{named}

\end{document}